\documentclass[sigconf]{acmart}
\usepackage{subfigure}
\usepackage{graphicx}
\usepackage{multirow}
\usepackage{tabularx}
\usepackage[ruled]{algorithm2e}
\usepackage{color}

\newcolumntype{C}[1]{>{\PreserveBackslash\centering}p{#1}}
\newcolumntype{R}[1]{>{\PreserveBackslash\raggedleft}p{#1}}
\newcolumntype{L}[1]{>{\PreserveBackslash\raggedright}p{#1}}

\AtBeginDocument{%
  \providecommand\BibTeX{{%
    \normalfont B\kern-0.5em{\scshape i\kern-0.25em b}\kern-0.8em\TeX}}}

\copyrightyear{2020}
\acmYear{2020}
\setcopyright{acmcopyright}\acmConference[MM '20]{Proceedings of the 28th ACM International Conference on Multimedia}{October 12--16, 2020}{Seattle, WA, USA}
\acmBooktitle{Proceedings of the 28th ACM International Conference on Multimedia (MM '20), October 12--16, 2020, Seattle, WA, USA}
\acmPrice{15.00}
\acmDOI{10.1145/3394171.3413765}
\acmISBN{978-1-4503-7988-5/20/10}

\begin{document}
\fancyhead{}

\title{Multi-modal Cooking Workflow Construction for Food Recipes}

\author{Liangming Pan}
\affiliation{%
  \institution{National University of Singapore}
  \city{Singapore}
  \state{Singapore}
}
\email{e0272310@u.nus.edu}

\author{Jingjing Chen}
\authornote{Corresponding Author}
\affiliation{%
  \institution{Fudan University}
  \city{Shanghai}
  \country{China}}
\email{chenjingjing@fudan.edu.cn}

\author{Jianlong Wu}
\affiliation{%
  \institution{Fudan University}
  \city{Shanghai}
  \country{China}}
\email{19210240015@fudan.edu.cn}

\author{Shaoteng Liu}
\affiliation{%
 \institution{Xi'an Jiaotong University}
 \city{Xi'an}
 \state{Shanxi}
 \country{China}}
\email{ls2662@stu.xjtu.edu.cn}

\author{Chong-Wah Ngo}
\affiliation{%
  \institution{City University of Hong Kong}
  \city{Hong Kong}
  \country{China}}
\email{cscwngo@cityu.edu.hk}

\author{Min-Yen Kan}
\affiliation{%
  \institution{National University of Singapore}
  \city{Singapore}
  \state{Singapore}
}
\email{kanmy@comp.nus.edu.sg}

\author{Yugang Jiang}
\affiliation{%
  \institution{Fudan University}
  \city{Shanghai}
  \country{China}}
\email{ygj@fudan.edu.cn}

\author{Tat-Seng Chua}
\affiliation{%
  \institution{National University of Singapore}
  \city{Singapore}
  \state{Singapore}
}
\email{dcscts@nus.edu.sg}

\renewcommand{\shortauthors}{Pan, et al.}

\begin{abstract}
Understanding food recipe requires anticipating the implicit causal effects of cooking actions, such that the recipe can be converted into a graph describing the temporal workflow of the recipe. This is a non-trivial task that involves common-sense reasoning. However, existing efforts rely on hand-crafted features to extract the workflow graph from recipes due to the lack of large-scale labeled datasets. Moreover, they fail to utilize the cooking images, which constitute an important part of food recipes. In this paper, we build \textit{MM-ReS}, the first large-scale dataset for cooking workflow construction, consisting of 9,850 recipes with human-labeled workflow graphs. Cooking steps are multi-modal, featuring both text instructions and cooking images. We then propose a neural encoder--decoder model that utilizes both visual and textual information to construct the cooking workflow, which achieved over 20\% performance gain over existing hand-crafted baselines. 
\end{abstract}



\maketitle

\section{Introduction}

\begin{figure*}[!t]
\centering
\subfigure[The cooking recipe of ``Blueberry Crumb Cake''. ]
{
	\begin{minipage}[t]{0.35\linewidth}
	\centering
	\includegraphics[width=6.0cm]{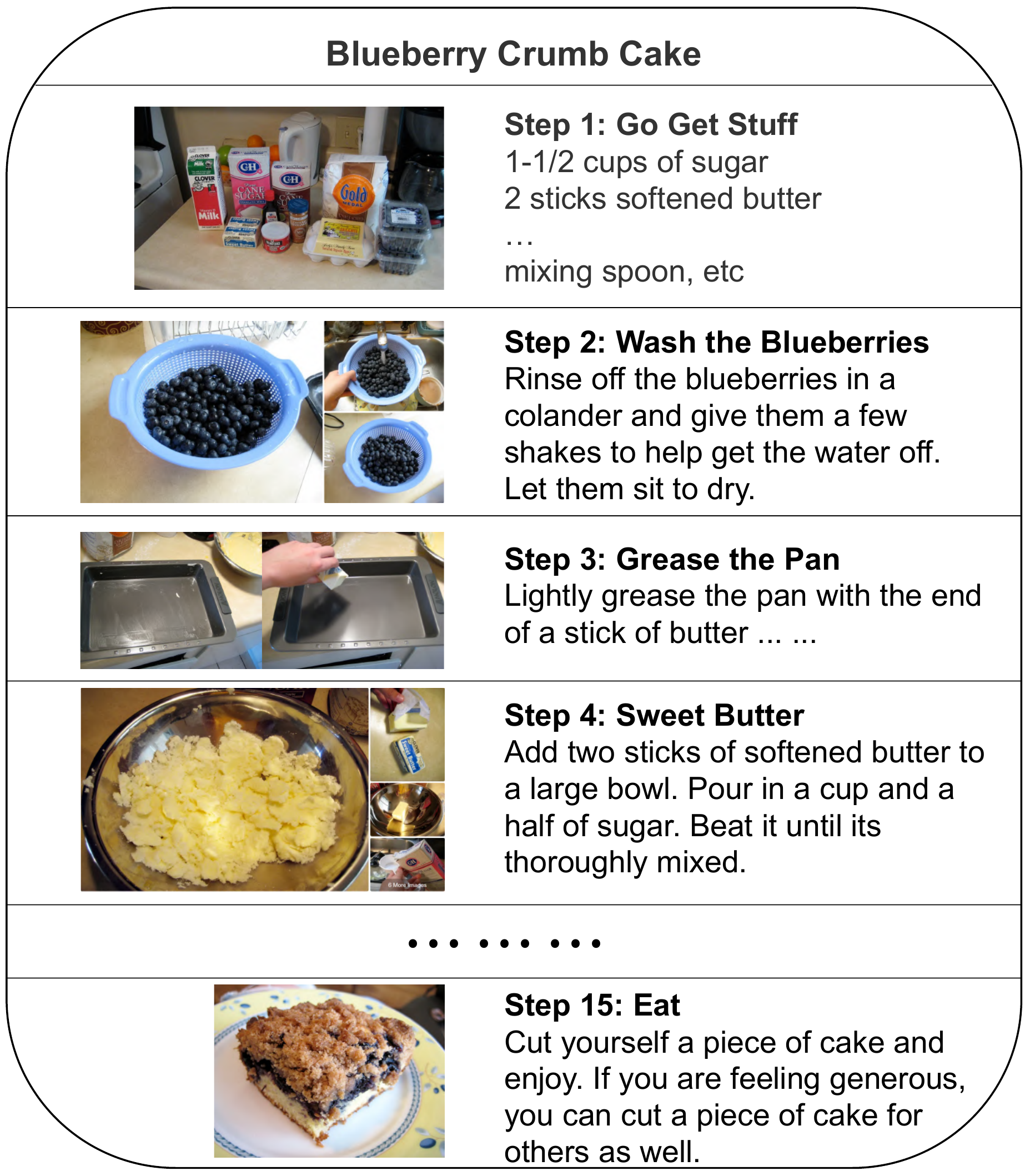}
	\end{minipage}
	\label{fig:recipe_example}
}
\subfigure[The cooking workflow for ``Blueberry Crumb Cake''. ]
{
	\begin{minipage}[t]{0.55\linewidth}
	\centering
	\includegraphics[width=10.0cm]{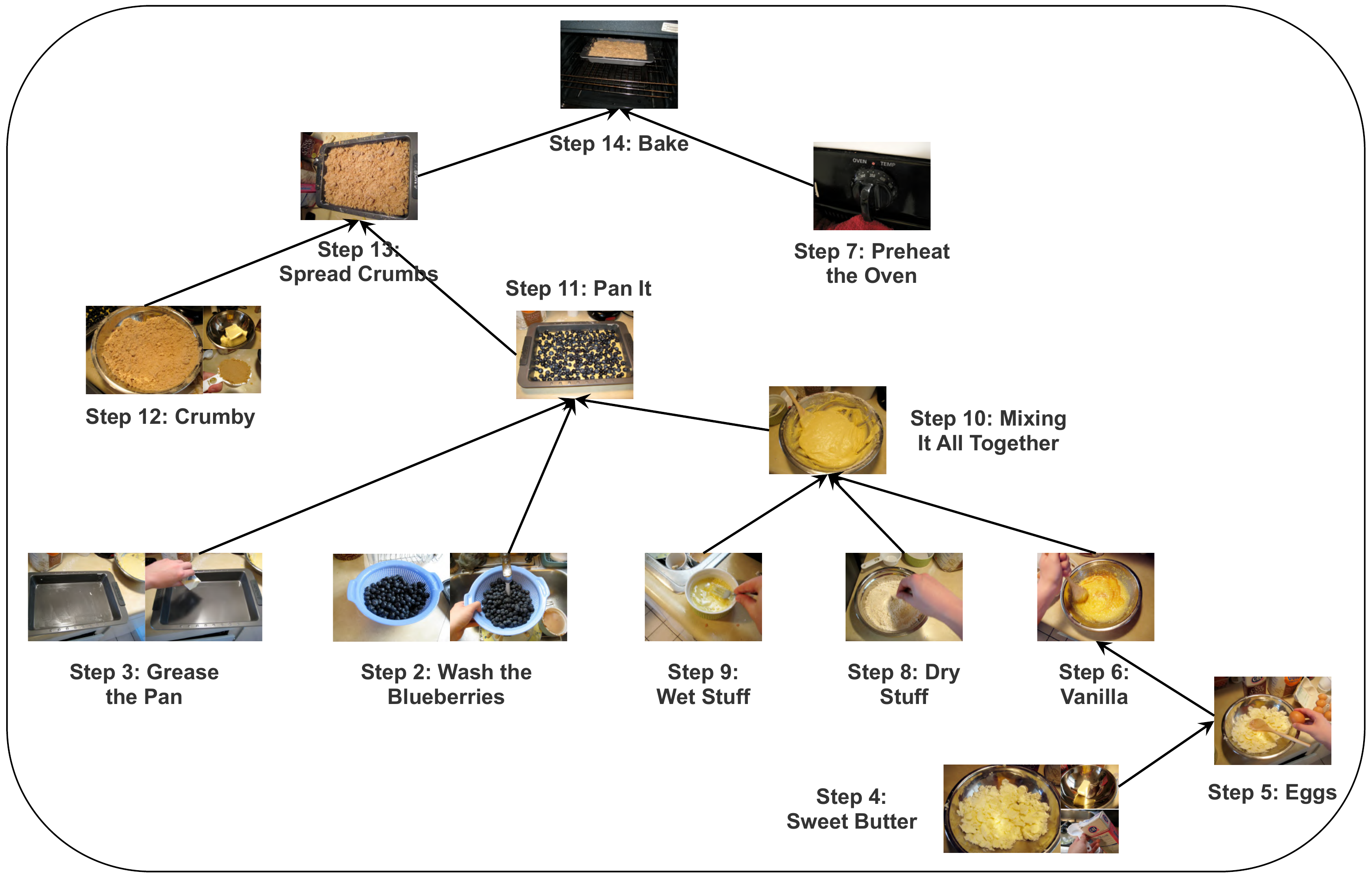}
	\end{minipage}
	\label{fig:workflow_example}
}
\caption{The cooking recipe ``Blueberry Crumb Cake'' (a) and its corresponding cooking workflow (b). }
\label{fig:task_example}
\end{figure*}

Nowadays, millions of cooking recipes are available online on cooking sharing platforms, such as AllRecipes, Cookpad, and Yummly, etc. A recipe is usually presented in multimedia setting, with textual description of cooking steps aligned with cooking images to illustrate the visual outcome of each step. See Figure~\ref{fig:recipe_example} for multimedia presentation of the recipe for ``Blueberry Crumb Cake''. These information potentially provide opportunity for multi-modal analysis of recipes, including cuisine classification~\cite{DBLP:journals/tmm/MinBMZRJ18}, food recognition~\cite{DBLP:journals/tmm/XuHJWSJ15,DBLP:journals/tmm/HerranzJX17}, recipe recommendation~\cite{maruyama2012real,DBLP:journals/tmm/HoriguchiAOA18} and cross-modal image-to-recipe search~\cite{salvador2017learning,chen2017cross,min2017being,carvalho2018cross}. A common fundamental problem among these tasks is in the modeling of the cause-and-effect relations of this \textbf{cooking workflow construction}. In this paper, we investigate this problem leveraging multiple modalities. 

\begin{figure}[t]
\centering
\subfigure[\textit{Sequential} relationship. ]
{
	\begin{minipage}[t]{1.0\linewidth}
	\centering
	\includegraphics[width=7.0cm]{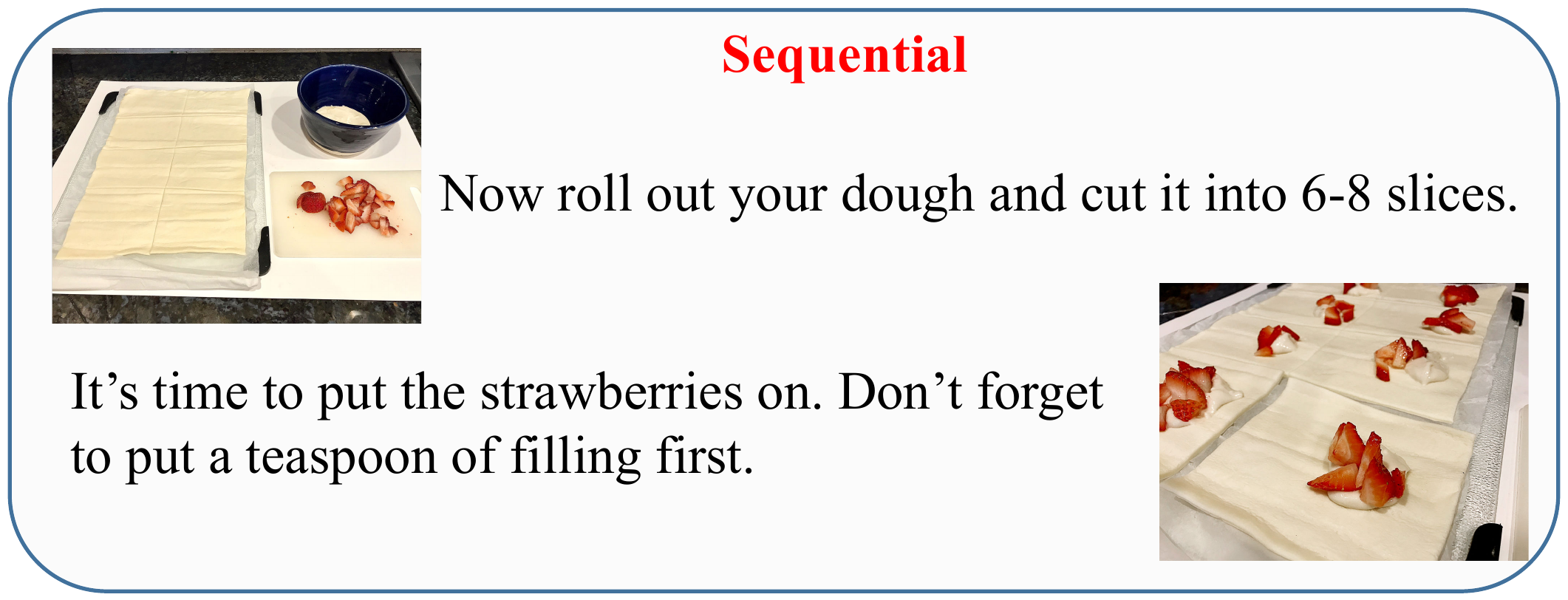}
	\end{minipage}
	\label{fig:exp_sequential}
}
\subfigure[\textit{Parallel} relationship. ]
{
	\begin{minipage}[t]{1.0\linewidth}
	\centering
	\includegraphics[width=7.0cm]{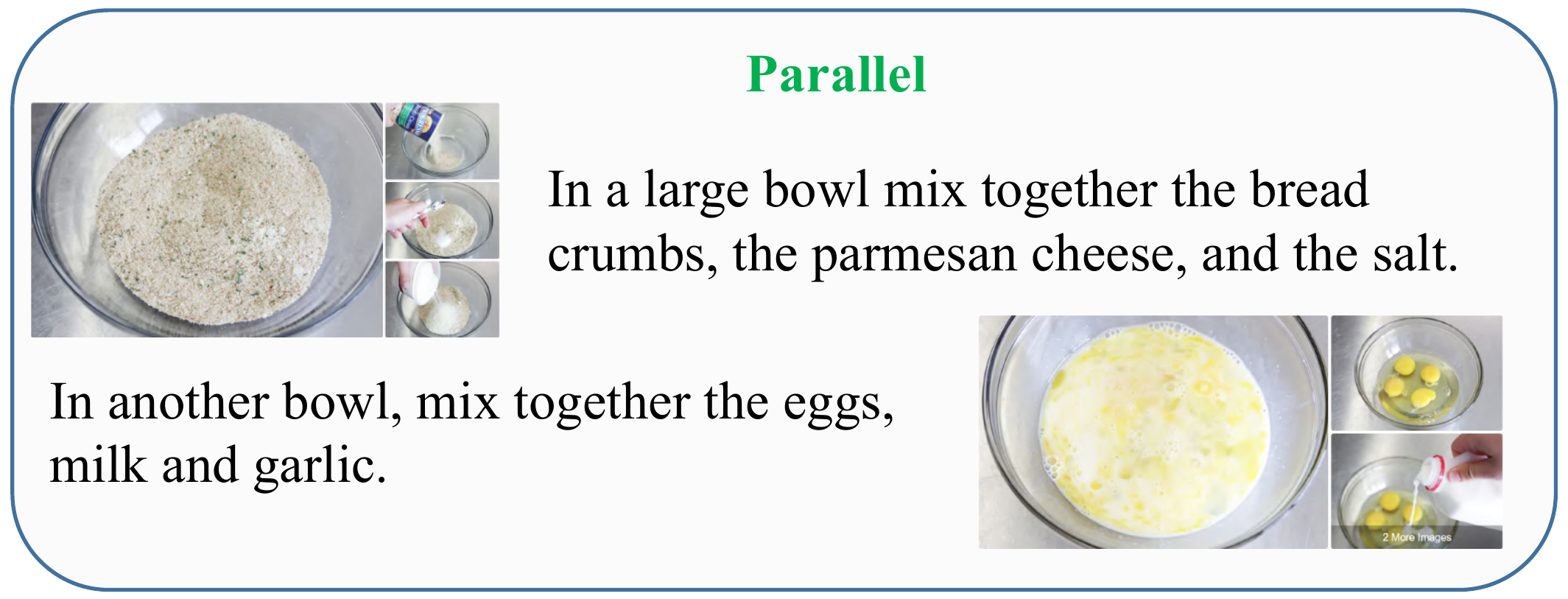}
	\end{minipage}
	\label{fig:exp_parallel}
}
\caption{Examples of sequential relationship (a) and parallel relationship (b) between two cooking steps. }
\label{fig:motivating_example}
\end{figure}

In food recipes, two cooking steps can be either \textit{sequential} or \textit{parallel}, as exemplified in Figure~\ref{fig:motivating_example}. Sequential means we cannot perform one step without finishing the other, while parallel indicates that the two steps are independent and can be performed at the same time. Based on these relations between pairs of cooking steps, we can draw a \textit{cooking workflow} that describes the temporal evolution of the food's preparation; see Figure~\ref{fig:workflow_example}. Formally, a cooking workflow is represented as a graph, where each node represents a cooking step. The nodes are chained in temporal order, where a link represents a sequential relation. 

The problem of cooking workflow construction has not been fully explored and mostly addressed with text-only analysis. For example, text-based dependency parsing is employed for workflow construction~\cite{kiddon2015mise,yamakata2016method,yamakata2017cooking}, and the hierarchical LSTM has been applied to model the causality effect for feature embedding~\cite{DBLP:conf/iclr/ChungAB17}. However, we believe this problem should be addressed with multi-modal analysis for two reasons. 

First, \textit{textual descriptions and cooking images usually play complementary roles in detecting cause-and-effect relations}. Figure~\ref{fig:exp_sequential} shows an example on why relying on text description alone is insufficient. In this example, we cannot infer the casual relation purely from the text description, as it does not mention where the strawberries are to be placed. But the pragmatic problem can be solved by looking at the cooking images of the two instructions. Similarly, the use images alone does not give sufficient clue to determine the casual relation between two steps. As shown in Figure~\ref{fig:exp_parallel}, although the cooking images look similar, the two steps are actually parallel, which can only be inferred from the clue ``in another bowl'' in the text description. 

Second, \textit{a multi-modal cooking workflow has wider applications in both real-world cooking and recipe-related research}. In real life, a workflow with both images and texts provides a more intuitive guidance for cooking learners. Novel recipe-based applications can also be proposed: in image-to-recipe generation, it is easier for machines to write a recipe following the guidance of a cooking workflow; in cross-modal retrieval, the model can benefit from the additional knowledge of cause-and-effect relations; in food recommendation, two recipes can be associated based on the structural similarity of their cooking workflows. 

Despite its importance, understanding the cause-and-effect relations in a cooking recipe is a non-trivial problem, usually requiring an in-depth understanding of both the visual and textual information. On the visual side, visually similar steps are not necessarily sequential, exemplified by Figure~\ref{fig:exp_parallel}. Therefore, fine-grained ingredient recognition is often required; \textit{e.g.}, the two steps in Figure~\ref{fig:exp_sequential} are sequential because they both operate on the strawberries. However, accurate ingredient recognition is quite challenging because of the variety in appearance of ingredients, resulting from various cooking and cutting methods~\cite{aaai20/chenzero}. On the textual side, understanding causal relations also requires an in-depth understanding of the cooking instruction as well as the contexts from previous steps. 

Neural networks, especially deep visual and language models such as ResNET~\cite{he2016deep} and BERT~\cite{DBLP:conf/naacl/DevlinCLT19}, offer promising solutions for the above challenges by learning deep visual and textual features. However, training them often requires a large amount of labeled data. Existing datasets, \textit{i.e.}, the Recipe Flow-graph Corpus (r-FG)~\cite{DBLP:conf/icmcs/YamakataIMM16} and the Carnegie Mellon University Recipe Database (CURD)~\cite{tasse2008sour} only have 208 and 260 labeled cooking workflows, respectively. Moreover, none of these datasets include cooking images. Due to the limited dataset scale, existing methods are largely restricted to using hand-crafted textual features, such as matching words~\cite{jermsurawong2015predicting}, and syntactic parsing~\cite{DBLP:conf/icmcs/YamakataIMM16}. These features are only able to capture shallow semantics, in addition to ignoring visual information.  

To address the above problems, we construct a large-scale dataset, namely the \textit{Multi-modal Recipe Structure dataset} (MM-ReS), consisting of 9,850 recipes with labeled cooking workflows. Each recipe contains an average of 11.26 cooking steps, where each step comprises of both textual instructions and multiple cooking images. We then propose a neural model which employs the Transformer architecture~\cite{DBLP:conf/nips/VaswaniSPUJGKP17} and the idea of Pointer Network~\cite{DBLP:conf/nips/VinyalsFJ15} to construct the cooking workflow. We compare our method with existing hand-crafted baseline~\cite{jermsurawong2015predicting} as well as strong neural baselines such as BERT~\cite{DBLP:conf/naacl/DevlinCLT19} and Multimodal Bitransformers (MMBT)~\cite{DBLP:conf/nips/KielaBFT19}. Experiment results show that neural-based models outperform hand-crafted baseline by a large margin. Neural models which utilize both recipe texts and cooking images generally perform better than models using a single modality. Among them, our proposed model achieves the best average $F_1$ score. To the best of our knowledge, this is the first work that explores multi-modal information for detecting cause-and-effect relationship in cooking recipes. 

\section{Related Work}
\label{sec:related_works}



\subsection{Cooking Workflow Construction}
\label{sec:recipe_structure_modeling}

Existing works~\cite{wang2008substructure, yamakata2017cooking, yamakata2016method, yamakata2013feature, hamada2005cooking, karikome2012improving, walter2011workflow}  for cooking workflow construction can be categorized into \textit{ingredient-level} and \textit{instruction-level} methods, based on the granularity of the workflow. 

Ingredient-level methods aim to parse a recipe into a work-flow graph, where each vertex represents either a cooking action or a raw ingredient, and directed edges represent the ``action flow'' (describing the temporal execution sequence) or ``ingredient flow'' (tracking the ingredient sources). Early work manually built the workflow graph for accurate recipe retrieval~\cite{wang2008substructure,xie2010hybrid, mori2014flow}, requiring laborious human labeling. An unsupervised hard-EM approach was proposed to automatically build workflow graphs by alternately optimizing a segmentation and a graph model \cite{kiddon2015mise}. The segmentation extracts actions from the text recipe while the graph model defined a distribution over the connections between actions. Yamakata \textit{et al.}~\cite{yamakata2016method} further proposed to enrich the workflow graph with cooking tools and duration with a semi-supervised method with four steps: word segmentation, recipe term identification, edge weight estimation, and manual action graph refinement. Nevertheless, ingredient-level methods do not attain high quality workflows for real-world applications due to two reasons: (1) the results are highly dependant on NLP tasks -- such as named entity recognition, co-reference resolution and dependency parsing -- which are noisy due to varied writing style in recipes, and (2) the lack of large-scale labeled fine-grained recipe structure data, also infeasible due to the required manual effort.

Compared with ingredient-level methods, instruction-level workflow is more practical in terms of scalability. In~\cite{jermsurawong2015predicting}, an ingredient-instruction dependency tree representation named SIMMER was proposed to represent the recipe structure. SIMMER represents a recipe as a dependency tree with ingredients as leaf nodes and recipe instructions as internal nodes. In SIMMER, several hand-crafted text features were designed to train the Linear SVM-rank model for predicting links between instructions. Similar to \cite{jermsurawong2015predicting}, we also focus on instruction-level workflow construction; however, we study from the perspective of multi-modal learning by considering both text procedures and cooking images in the recipe. Moreover, instead of defining hand-crafted features, we improve the feature extraction using neural models to obtain deep semantic features. 

\subsection{Prerequisite Relation Detection}

The key to building a cooking workflow lies in detecting the parallel/sequential relationship, which is essentially a kind of prerequisite relation. Despite being a relatively new research area, data-driven methods for learning concept prerequisite relations have been explored in multiple domains. In educational data mining, prerequisite relations have been studied among courses or course concepts for curriculum planning \cite{pan2017prerequisite, liang2017recovering, yang2015concept, liu2011mining}. Pan \textit{et al.}~\cite{pan2017prerequisite,DBLP:conf/ijcnlp/PanWLLT17} proposed hand-crafted features such as video references and sentence references for learning prerequisite relations among concepts in MOOCs. Besides education domain, prerequisite relation has also been mined between Wikipedia articles \cite{liang2015measuring, talukdar2012crowdsourced}, concepts in textbooks \cite{wang2016using}, as well as concepts in scientific corpus \cite{gordon2016modeling}. 

Existing methods are limited to hand-crafted textual features, such as the maximum matching words~\cite{jermsurawong2015predicting}, reference distance~\cite{liang2015measuring}, and complexity level distance~\cite{pan2017prerequisite}. Although these features capture shallow semantics, they are mostly domain-dependent and not transferable across applications. Furthermore, as existing work has focused only on pure text, such as Wikipedia page and textbooks.  How to best make use of the multimedia nature of documents in describing causality has been insufficiently investigated.

\subsection{Cross-modal Food Analysis}

Cross-modal learning in food domain has started to attract research interest in recent years. Novel tasks such as ingredient/food recognition~\cite{DBLP:conf/aaai/ChenPWWNC20}, cross-modal retrieval~\cite{DBLP:conf/sigir/CarvalhoCPSTC18,DBLP:conf/cvpr/WangSLLH19} and recipe generation~\cite{DBLP:conf/cvpr/SalvadorDNR19} have been proposed, and several large food and recipe datasets have been developed; for example, Cookpad~\cite{DBLP:conf/sigir/HarashimaSK17} and Recipe1M+~\cite{DBLP:journals/corr/abs-1810-06553}. Existing neural-based methods~\cite{min2017being,DBLP:conf/cvpr/SalvadorHAMOW017,salvador2017learning,liu2020hyperbolic} typically learn a joint embedding space between food images and recipe texts. 
For example, in~\cite{min2017being}, a deep belief network is used to learn the joint space between food images and ingredients extracted from recipes. 
However, previous works consider a recipe as a whole, but ignore its inherent structure. Different from these works, our work investigate the cause-and-effect relations inherent in cooking recipes, based on which we can learn better recipe representations to benefit downstream tasks. 


\section{Dataset: MM-ReS}

Cooking workflow construction is a novel task that lacks a large-scale dataset. To facilitate future research, we construct the \textit{Multi-modal Recipe Structure} (MM-ReS) dataset, containing 9,850 real food recipes. MM-ReS is the first large scale dataset to simultaneously contain: (1) labeled cooking workflow for each food recipe, and (2) cooking images and text descriptions for each cooking step. 

\subsection{Data Collection}

We collect food recipes from two cooking sharing platforms: Instructables\footnote{https://www.instructables.com/} and AllRecipes\footnote{https://www.allrecipes.com/} (statistics summarized in Table~\ref{tbl:raw_data}):

\noindent $\bullet$ \textbf{Instructables} is one of the largest do-it-yourself (DIY) sharing platforms, which contains millions of user-uploaded DIY projects, including over 30,000 food recipes. Users post step-by-step cooking instructions to the recipe, with each step accompanied by one or multiple cooking images (see Figure~\ref{fig:recipe_example} as an example). We crawled all recipes under the category ``food'' and excluded none-English recipes, resulting in a total of 32,733 recipes. On average, each crawled recipe contains 5.65 cooking steps while each step contains 2.32 cooking images. As the recipes are written by contributing users, the texts are relatively noisy and include information irrelevant to the cooking procedure, such as ``Look, we are done, excited?''. The cooking steps divided by users are often in \textit{coarse-grained}, with each step containing multiple cooking actions. 

\noindent $\bullet$ \textbf{AllRecipes} is an advertising-based revenue generator, presented as a food focused online social networking service. The recipes on the website are posted by members of the Allrecipes.com community. They are categorized by season, type (such as appetizer or dessert), and ingredients. We crawled all English recipes from the website and obtain 65,599 valid recipes. Compared with the recipes from Instructables, the recipes in AllRecipes are written by experts, therefore the texts are of \textit{high quality} and the cooking steps are more \textit{fine-grained}, with each step only corresponds to one or two cooking actions. Despite with high quality, recipes do not have cooking images associated with each step. However, a portion of recipes have high-quality cooking videos made by the website, serving as a good source to extract cooking images.  

\begin{table}
  \caption{Basic statistics of collected recipes}
  \label{tbl:raw_data}
  \begin{tabular}{ccc}
  \hline
    Data Source & Instructables & AllRecipes\\ \hline
    \# Recipes & 32,733 & 64,500 \\
    \# Cooking Images & 184,941 & --- \\
    \# Sentences & 161,046 & 120,615 \\
    Text Quality & Noisy & Clean \\
    Cooking Steps & Coarse-grained & Fine-grained \\ \hline
    \end{tabular}
\end{table}

\subsection{Data Processing}

We first process the collected data in two steps:

1) \textbf{Data Filtering. }
We first select high-quality recipes from the collected data to construct our final dataset. For recipes from Instructables, we discard recipes that contain less than $7$ steps\footnote{We exclude the initial steps that introduce ingredients.} as their cooking workflows are likely to form trivial chains, rather than a graph structure. We also ensure that each step has both a text description and at least one cooking image.  User-contributed cooking steps are often lengthy, describing multiple cooking actions.  We split steps consisting of more than 3 sentences into individual sentences, treating each as one cooking step. We obtain $5,071$ high-quality recipes from Instructables after data filtering. For recipes from AllRecipes, the cooking steps are already fine-grained. 
We then rank the recipes by the number of cooking steps, and selecting the first $5,000$ recipes that have well-made cooking videos. 

2) \textbf{Key Frame Extraction.} To obtain cooking images for recipes in AllRecipes, we extract key frames from cooking videos. We first extract frames from each recipe video with fixed time intervals using the  \textit{ffmpeg}\footnote{https://www.ffmpeg.org/} video processing toolkit. We then select key frames by filtering out images that are similar or with low resolution. Specifically, we extract visual features for each candidate frame using pre-trained ResNet-50~\cite{he2016deep}. If the cosine similarity between two consecutive frames are above a certain threshold, we only keep one frame and delete the other. In the end, we obtain 131,135 cooking images (an average of 26.23 images for a recipe). 

\subsection{Alignment between Text and Image}

We then align each cooking step with its cooking images. For recipes from Instructables, because each long step has been split into multiple mini-steps, we need to assign cooking images for each mini-step. For AllRecipes, the cooking images extracted from cooking video are not assigned to each cooking step in the recipe. Therefore, we ask human annotators to align cooking images to their corresponding cooking steps. Specifically, we hire 16 undergraduates who are native English speakers with cooking experience as annotators. We build an annotation platform in which the alignment task is formulated in the form of multiple-choice. For each step, we show its text description at the top, and its candidate cooking images below. The annotator is required to choose the image(s) that matches the text description, and choose ``No Picture Present That is Related'' if there is no image can be matched. Moreover, we also filter out irrelevant cooking steps in this process: if the text description is not related to cooking, the annotator is required to choose the ``Sentence Not Related To Cooking Procedure''. In total, 227,082 cooking images are aligned to 110,878 cooking steps, with 10,000 steps are doubly annotated to determine the inter-annotator agreement. The whole annotation process takes 2 months, costing 240 man-hours. The inter-annotator agreement reached a Cohen’s~Kappa of 0.82, suggesting a substantial agreement. 


\begin{figure}[!t]
\centering
	\includegraphics[width=8.0cm]{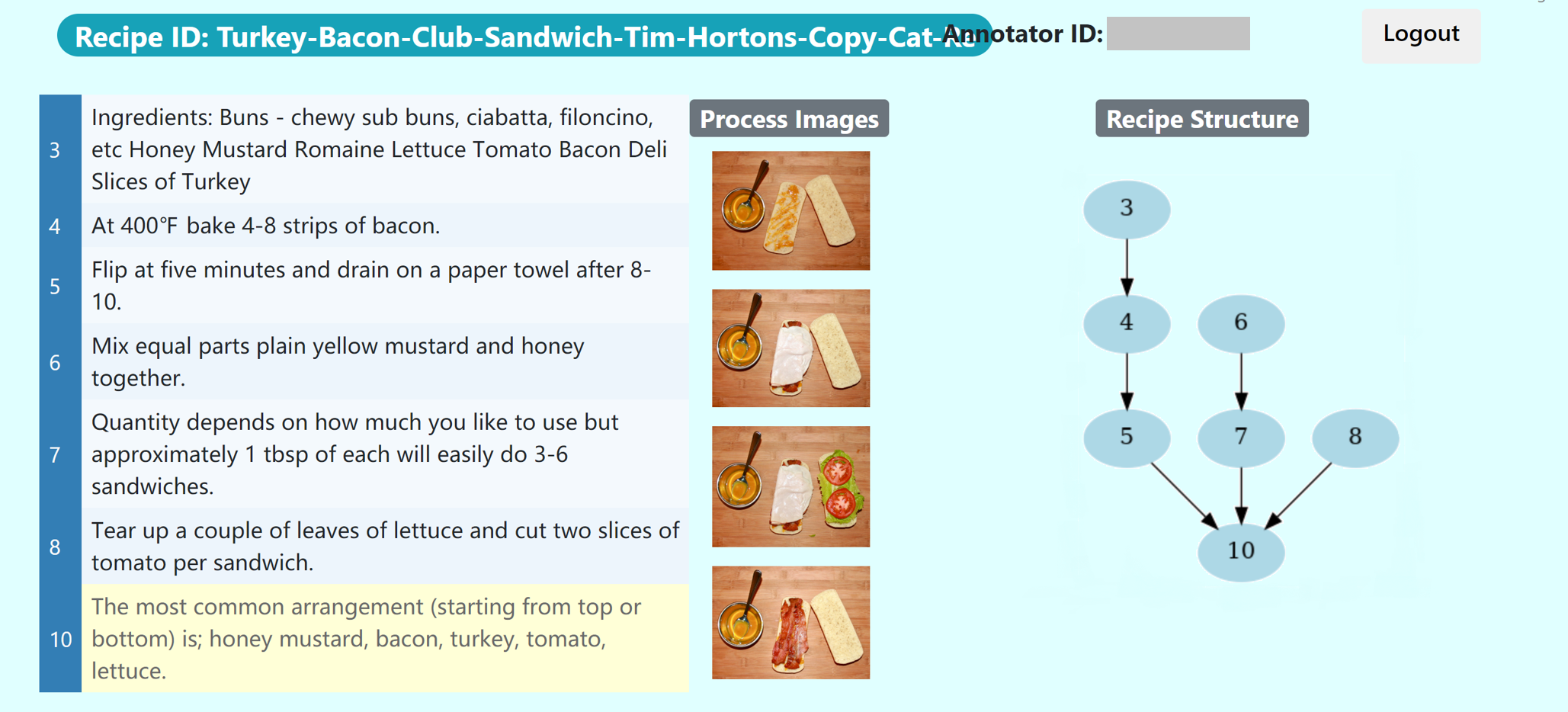}
\caption{Our cooking workflow annotation platform.}
\label{fig:annotation_demo}
\end{figure}

\begin{table*}[!t]
  \caption{The recipe-, image-, step-, and sentence-level data statistics of the MM-ReS dataset}
  \label{tab:statistics}
  \begin{tabular}{clc || clc} \hline
    Type & Features & Number  & Type & Features & Number \\ \hline
    \multirow{7}{*}{Recipe} & Number & 9,850 & \multirow{7}{*}{Step} & Number & 110,878 \\
    & \# Recipes from Instructables & 5,013 (50.9\%) & & \# Cooking Steps & 81,615 (73.6\%) \\
    & \# Recipes from AllRecipes & 4,837 (49.1\%) & & \# Non-Cooking Steps & 29,263 (26.4\%) \\
    & \# Avg. Steps / Recipe & 11.26 & & \# Cooking Steps with Images & 65,969 (80.83\%) \\
    & \# Avg. Cooking Steps / Recipe & 8.29 & & \# Avg. Sentences / Step & 1.29  \\
    & \# Avg. Tokens / Recipe & 228.8 & & \# Avg. Tokens / Step & 20.33 \\
    & \# Avg. Images / Recipe & 23.05 & & \# Avg. Images / Step & 2.05 \\ \hline
    \multirow{2}{*}{Image} & Number & 227,082 & \multirow{2}{*}{Sentence} & Number & 143,580 \\
    & \# Images linked to recipe & 179,975 (79.25\%) & & \# Avg. Tokens / Sentence & 15.70 \\ \hline
\end{tabular}
\end{table*}

\subsection{Cooking Workflow Construction}

After each cooking step is aligned with its cooking images, we then construct the cooking workflow for each of the 10,071 recipes (5071 from Instructables; 5000 from AllRecipes) obtained after data processing. We hire 22 English-speaking undergraduate students with cooking experience to manually annotate the cooking flow. To facilitate the annotation, we build an annotation platform as shown in Figure~\ref{fig:annotation_demo}. The recipe is shown on the left table, with each row being a cooking step. Note that we filter out the steps labeled as irrelevant in the text-image alignment process. When the annotator moves over a certain step, its cooking images are shown in the middle. The cooking workflow for the recipe is shown on the right, in which each node represents a step in the left recipe. Initially, the cooking workflow is empty with no link between nodes. The annotator is required to construct the workflow by chaining the nodes based on relations, where sequential relation results in a link between two nodes. The annotation takes 1 months, costing 310 man-hours. We hire two students expertise in cooking to run a quality control over the annotated recipes, filtering out 221 low-quality annotations. Among the 9,850 valid recipes, 1,500 recipes are randomly sampled for double annotation to determine the inter-annotator agreement. We covert each annotation as an one-hot vector of all possible node pairs, based which the Cohen’s Kappa is calculated as 0.71, suggesting a substantial agreement. 


\subsection{Data Statistics}

The MM-ReS dataset contains 9,850 recipes, 110,878 steps, and 179,975 aligned cooking images. Detailed data statistics are in Table~\ref{tab:statistics}. 
The MM-ReS dataset has two distinct features compared with other existing food datasets. First, it is the first food dataset that has multi-modal information on step-level, with each cooking step associated with both texts and images. Existing food datasets either only have text (\textit{e.g.}, YOUCOOK2~\cite{DBLP:conf/aaai/ZhouXC18}) or images (\textit{e.g.}, Food-101~\cite{DBLP:conf/eccv/BossardGG14}), or the cooking image is on the recipe-level rather than step-level (\textit{e.g.}, Recipe1M+~\cite{marin2019learning}). Second, our dataset contains $9,850$ human-annotated cooking workflows; this scale exceeds other datasets with recipe workflows by almost two magnitudes, such as r-FG~\cite{DBLP:conf/icmcs/YamakataIMM16} and the CURD~\cite{tasse2008sour}. 

\section{Methodology}
\label{sec:method}

We first formally define the problem of cooking workflow construction, then introduce our proposed model. 

\subsection{Problem Formulation}
\label{sec:problem_define}

A \textbf{recipe} $R$ is composed of $n$ cooking steps, denoted as $R = \{ S_1, \cdots, S_n \}$, where $S_i$ is the $i$-th step. Each \textbf{cooking step} $S$ is further represented as its text description and cooking images, \textit{i.e.}, $S = \{ \mathcal{W}, \mathcal{I} \}$, where the text description $\mathcal{W} = ( w_1, \cdots, w_{|\mathcal{W}|} )$ is a word sequence, and $\mathcal{I} = \{ x_1, \cdots, x_{|\mathcal{I}|} \}$ is a set of cooking images. 
The \textbf{cooking workflow} of a recipe $R$ is defined as a directed graph $G = (\mathcal{V}, \mathcal{E})$, where each cooking step $S_i$ is represented as a vertex in $\mathcal{V}$. A directed edge $e = \langle S_i, S_j \rangle$ from $S_i$ to $S_j$ exists if: 
\begin{enumerate}
    \item $i < j$, \textit{i.e.}, step $S_i$ appears before $S_j$ in the recipe. 
    \item $S_i$ and $S_j$ have a causal dependency, \textit{i.e.}, we cannot perform step $S_j$ without finishing step $S_i$. 
\end{enumerate}
Figure~\ref{fig:workflow_example} shows an example of cooking workflow. Step $3$ is a prerequisite step of $11$ since the pan has to be prepared before adding blueberries into it. However, step $8$ and step $9$ can be processed in parallel since the dry ingredients and wet ingredients can be prepared independently. By following the edges, we can clearly tell how the food can be prepared in an efficient and collaborative way. 

Given a recipe $R$ as input, the objective is to build the cooking workflow $G$. The major challenge lays in how to judge whether two steps have a causal dependency. We address this by extracting deep semantic features from both images and texts and proposing a neural model based on the Transformer~\cite{DBLP:conf/nips/VaswaniSPUJGKP17} and the Pointer Network~\cite{DBLP:conf/nips/VinyalsFJ15} to detect causal relations. 


\subsection{Model Framework}
\label{sec:model_framework}

\begin{figure*}[!t]
\centering
	\includegraphics[width=16.0cm]{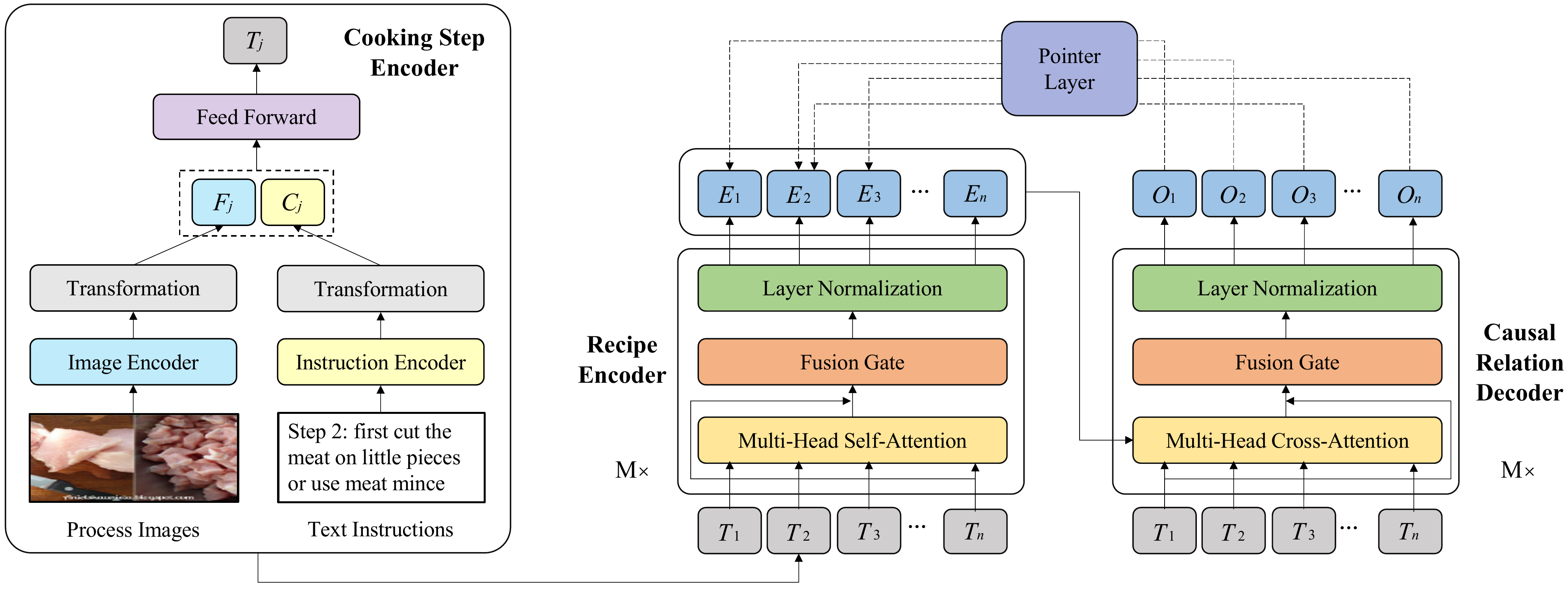}
\caption{The general framework of the proposed model for cooking workflow construction. }
\label{fig:model_framework}
\vspace{-0.2cm}
\end{figure*}

The $i$-th cooking step is $S_i = \left( \{ w_{i, 1}, w_{i, 2}, \cdots, w_{i, L_i} \}, \{ x_{i, 1}, \cdots, x_{i, M_i} \} \right )$, 
where $L_i$ is the number of words in the text description ($w_{i,j}$ denotes the $j$-th word), and $x_{i,k}$ is the $k$-th cooking image associated with the step. For each step $S_i$, the goal of our model is to predict $p(S_j \vert S_i)$, \textit{i.e.}, the probability that $S_j$ is a prerequisite step of $S_i$. Given a training set of $N$ recipe-workflow pairs $\{ (R^{(i)}, G^{(i)}) \}_{i=1}^N$, our model is trained to maximize the following likelihood function: 


\begin{equation}
    \label{equ:train_obj}
    \sum_{i=1}^N \frac{1}{|\mathcal{E}^{(i)}|} \sum_{\langle S_j, S_k \rangle \in \mathcal{E}^{(i)}} \log p(S_k \vert S_j)
\end{equation}

\noindent where $\mathcal{E}^{(i)}$ is the set of edges in the workflow graph $G^{(i)}$. 

Our model is designed as an encoder--decoder architecture, composed of a recipe encoder and a relation decoder. The \textbf{recipe encoder} is a hierarchical structure. First, a \textit{cooking step encoder} is trained to learn the vector representation of a cooking step by integrating the information from text descriptions and cooking images (Section~\ref{sec:step_encoder}). The step embeddings are then fed into a transformer-based \textit{recipe encoder} for capturing global dependencies between steps (Section~\ref{sec:recipe_encoder}). Finally, the \textbf{relation decoder} utilizes the information captured by the recipe encoder to predict the prerequisite steps for each step one by one using a pointer network (Section~\ref{sec:decoder}). Figure~\ref{fig:model_framework} shows the overall architecture of our model. 

\subsection{Cooking Step Encoder}
\label{sec:step_encoder}

The cooking step encoder consists of an image encoder and an instruction encoder to learn embeddings for cooking images and instruction texts, respectively. The visual and textual embeddings are then fused to obtain the embedding for the cooking step. 
\\

\noindent \textbf{Image Encoder. }We use pre-trained ResNET-50~\cite{he2016deep} to extract features for cooking images. To make the model more adaptable to the food domain, we fine-tune the pre-trained ResNet-50 with Recipe1M~\cite{salvador2017learning} dataset, which contains $251,980$ training images of $1,047$ food categories (\textit{e.g.}, chocolate cake, cookie). During fine-tuning, the image features of ResNET-50 are projected to a softmax output layer to predict the food category during training. After fine-tuning, we drop the softmax layer and use the outputs from last layer as image features. Given the cooking images $\{ x_{i, 1}, \cdots, x_{i, M_i} \}$ for step $S_i$, the extracted visual feature for $x_{i, j}$ is denoted as $f_{i,j}$. 
The image encoder takes the average of $f_{i,1}, \cdots, f_{i, M_i}$ as the visual embedding, denoted as $F_i$. \\

\noindent \textbf{Instruction Encoder. }Given the text description of step $S_i$, denoted as a word sequence $\{ w_{i, 1}, w_{i, 2}, \cdots, w_{i, L_i} \}$, we use the pretrained GLoVE~\cite{DBLP:conf/emnlp/PenningtonSM14} as the word embeddings and employ a bidirectional LSTM~\cite{DBLP:journals/neco/HochreiterS97} to encode contextual information for words from both directions. We then aggregate the LSTM hidden states $h_{i,1}, \cdots, h_{i,T_i}$ into a single vector $C_i$ to represent the instruction text. Observing that some keywords like ``another'' and ``set aside'' may provide clues for casual relations, we obtain $C_i$ by applying a self-attention layer~\cite{DBLP:conf/nips/VaswaniSPUJGKP17} and aggregating the hidden states based on the learned attention weights. This endows the encoder with the ability to pay more attention to useful word clues. The vector $C_i$ is regarded as the semantic embedding of the instruction text. \\

\noindent \textbf{Multi-Modal Fusion. }We then propose the following two methods to fuse the image embedding $F_i$ and the instruction embedding $C_i$. 

1) \textbf{Concatenation. }
We apply a linear transformation separately for $F_i$ and $C_i$, and then we concatenate the two transformed vectors and apply a two-layer feed-forward to obtain the final step embedding, denoted as $T_i$. 

2) \textbf{MMBT. }
The concatenation-based method does not consider the interactions between $F_i$ and $C_i$. To address this, we employ the Multimodal Bitransformer model (MMBT)~\cite{DBLP:conf/nips/KielaBFT19} to capture high-order visual--textual interactions. MMBT feeds the text description and the cooking images together into a Transformer~\cite{DBLP:conf/nips/VaswaniSPUJGKP17} encoder to obtain the fused step embedding $T_i$. We use the pretrained version of MMBT\footnote{https://github.com/facebookresearch/mmbt/} and fine-tune it during training. 

\subsection{Recipe Encoder}
\label{sec:recipe_encoder}

Intuitively, judging the causal relation between two steps $S_i$ and $S_j$ also requires understanding the context information around $S_i$ and $S_j$. To this end, we input the $n$ step embeddings $\{T_1, \cdots, T_n\}$ into a \textit{recipe encoder}, which is composed of a stack of $M$ transformer layers~\cite{DBLP:conf/nips/VaswaniSPUJGKP17}. This outputs a set of \textit{contextualized} step embeddings $\{ E_1, \cdots, E_n \}$, where each step embedding $E_t$ not only encode the information of the step $S_t$, but also contains information of its context steps $S_1, \cdots, S_{t-1}, S_{t+1}, \cdots S_T$. 

Each transformer layer has three sub-layers. Multi-head self-attention mechanism is used for capturing the global dependencies between the $n$ steps in the recipe. A fusion gate is used to combine the input and output of the attention layer, which yields a self-aware and global-aware vector representation for each step. Finally, layer normalization~\cite{DBLP:journals/corr/abs-1805-07389} is implemented as the last part of the layer. 


\subsection{Causal Relation Decoder}
\label{sec:decoder}

The causal relation decoder sequentially predicts the causal relations for each step from $S_1$ to $S_n$. At time step $t$, the decoder takes the embedding of the current step $S_t$ as input, and it outputs a probability distribution over its previous steps $\{ S_1, \cdots, S_{t-1} \}$, denoted as $P_t$, where $P_{t, k}$ represents the probability for the step $S_k$ being a prerequisite step of $S_i$. 

The causal relation decoder is composed of a stack of $M$ transformer layers, plus a pointer layer over the final output of the decoder stack. Specifically, each transformer decoder layer consists of three parts: the multi-head cross-attention mechanism, a fusion gate, and a normalization layer. The query for the multi-head attention is the input embedding for the current step $t$, and the keys and values are the encoder outputs $\{ E_1, \cdots, E_n \}$. This allows the decoder to attend over all the steps $S_1, \cdots, S_n$ in the recipe to gather relevant information to predict the causal relation of the current step $t$; in other words, having a global understanding of the recipe contexts in decision making. 
These attention outputs are then fed to a fusion gate, followed by a normalization layer, similar to the recipe encoder. We denote the final output vector from the last decoder layer at time step $t$ as $O_t$. Finally, we stack a pointer layer for predicting the probability distribution $P_t$ based on the decoder output $O_t$, which is formulated as follows:
\begin{equation}
\centering
\label{equ:pointerlayer}
    Q_t = \text{ReLU}(O_t W_Q); \quad \quad K = \text{ReLU}(E_{1:t-1} W_K)
\end{equation}
\begin{equation}
\centering
\label{equ:pointerlayer2}
    P_t = \text{Sigmoid}(\frac{Q_t K^T}{\sqrt{d}})
\end{equation}

\noindent where $P_t$ is the output of the pointer layer, in which $P_{t,k}$ represents the probability that a causal relationship exists from $S_k$ to $S_t$. $W_Q$ and $W_K$ are parameter matrices. Note that we use the sigmoid function rather than softmax in the pointer layer as each step can be linked to multiple previous steps, therefore each $P_{t,k}$ is an independent probability between $0$ and $1$. 

After decoding, we obtain a set of model predictions $P = \{P_{i,j} \vert 1 \leq j < i \leq n \}$. We then use the following steps to construct the workflow graph. First, we select all edges $\langle S_j, S_i \rangle$ satisfying $P_{i, j}>\theta$ as candidate edges, where $\theta$ is a pre-defined threshold and is set to $0.5$ in the experiment. Second, we prune the graph by removing all redundant edges. Specifically, we remove the direct edge $\langle S_j, S_i \rangle$ if there exists a longer path $S_j \rightarrow \cdots S_k \cdots \rightarrow S_i$ from $S_j$ to $S_i$, because $S_j \rightarrow S_i$ is implied in this longer path. 



\section{Experiments}
\label{sec:experiments}

\subsection{Data and Metrics}
\label{sec:data_metrics}

We evaluate the performance of our method for cooking workflow construction on the MM-ReS dataset. The $9,850$ recipes in the dataset are randomly split into training (80\%), validation (10\%), and testing set (10\%). We report the performance on the testing set and tune the model parameters on the validation set. To evaluate the quality of the output workflow graph, we use the precision, recall, and $F_1$ score of predicting edges with respect to the ground-truth edges. The overall accuracy of the task is computed at the \textit{edge level} (counting all edges in the data set), and at the \textit{recipe level} (average accuracy over all recipes). Denote the ground-truth / predicted edge set for the $i$-th recipe in the test set as $E^{(i)}$ and $\hat E^{(i)}$, the edge-level precision ($P_{e}$) / recall ($R_{e}$) and the recipe-level precision ($P_r$) /recall ($R_r$) are calculated as follows:
\begin{equation}
\centering
\label{equ:evaluation_metrics}
    \begin{split}
        P_{e} = \frac{\sum_{i=1}^N \sum_{e \in |\hat E^{(i)}|} \mathbb{I}(e \in E^{(i)})}{\sum_{i=1}^N | \hat E^{(i)} |} ; \ 
        R_{e} = \frac{\sum_{i=1}^N \sum_{e \in |E^{(i)}|} \mathbb{I}(e \in \hat E^{(i)})}{\sum_{i=1}^N | E^{(i)} |} \\
        P_{r} = \sum_{i=1}^N \frac{\sum_{e \in |\hat E^{(i)}|} \mathbb{I}(e \in E^{(i)})}{| \hat E^{(i)} |} ; \
        R_{r} = \sum_{i=1}^N \frac{\sum_{e \in |E^{(i)}|} \mathbb{I}(e \in \hat E^{(i)})}{| E^{(i)} |}
    \end{split}
\end{equation}

\noindent $\mathbb{I}(s)$ is an indicator function, returning $1$ if the condition $s$ is true. 

\subsection{Baselines}
\label{sec:baseline}

We conduct a comprehensive performance evaluation for the following 10 methods, which can be categorized into three groups based on the use of information. 

\subsubsection{Textual-Only} We first choose three baselines that only utilize instruction texts in the recipe to build cooking workflow. 

\noindent $\bullet$ \textbf{Hand-crafted Features.} The work of~\cite{jermsurawong2015predicting} proposed several hand-crafted textual features, such as TF-IDF, to detect causal relations between two recipe instructions and train an SVM for relation classification. The predicted pairwise relations are used as the edges of the workflow graph. 

\noindent $\bullet$ \textbf{BERT Pairwise Detector. }To evaluate the effectiveness of deep textual features, we propose a baseline that applies BERT~\cite{DBLP:conf/naacl/DevlinCLT19} for pairwise causal relation detection. Specifically, we use BERT in double-sentence mode, in which the texts from step $S_i$ and $S_j$ are concatenated as a single sequence separated by a special token $\text{[SEP]}$. After taking this concatenated sequence as input, the output vector of the BERT is then linked to a feed forward network with $2$ hidden layers to predict casual relations. 

\noindent $\bullet$ \textbf{Ours (Instruction Encoder). }For our ablation study, we also employ a variant of our model that only has the instruction encoder when encoding a step. In this way, we ignore the cooking images by removing the image encoder. 


\subsubsection{Image-Only} We then include 3 baselines that detect causal relations purely based on the cooking images. 

\noindent $\bullet$ \textbf{Image Similarity Detector. }This is a weak baseline that judges causal relations based on image similarity, defined as the normalized cosine distance between ResNet-50 visual features. As a step is associated with multiple cooking images, we calculate the average / maximum / minimum image similarity between the cooking images from two steps. An SVM classifier is trained to learn the weights of these three similarities for relation detection.  

\noindent $\bullet$ \textbf{Feed-forward Neural Detector. }We adopt a two-layer feed forward neural network as a baseline for image-based relation detector. It takes as input the concatenation of the ResNet-50 visual features from two steps, and outputs a binary classification on whether the two steps are in sequential. 

\noindent $\bullet$ \textbf{Ours (Image Encoder). }This is a variant of our model that only has the image encoder when encoding a step. 

\subsubsection{Multi-Modal} Finally, we compare the following four methods that utilize multi-modal information of both images and texts. 

\noindent $\bullet$ \textbf{Multi-modal Hand-crafted Features. }We create a baseline that enriches the feature set of~\cite{jermsurawong2015predicting} by adding the three image similarity features described in the baseline Image Similarity Detector. 

\noindent $\bullet$ \textbf{MMBT Pairwise Detector.} This baseline applies MMBT~\cite{DBLP:conf/nips/KielaBFT19} for pairwise causal relation detection. Specifically, we concatenate the texts and images from step $S_i$ and $S_j$ as a sequence of embeddings, which is taken as inputs of the MMBT model for predicting the casual relation between $S_i$ and $S_j$. The MMBT model is trained by sampling the same number of sequential/parallel step pairs from the training set as positive/negative examples. 

\noindent $\bullet$ \textbf{Ours (Concatenation)}. This is  our full model, using vector concatenation to fuse the visual and textual embeddings. 

\noindent $\bullet$ \textbf{Ours (MMBT)}. This is our full model, using MMBT to fuse the visual and textual embeddings. 

\begin{table*}[!t]
\centering
\caption{Performance comparison with baselines and the ablation study. The best performance is in bold.}
\begin{tabular}{c l|c|c|c|c|c|c||c|c|c}
\hline
\multicolumn{2}{c|}{\multirow{2}{*}{Models}} & \multicolumn{3}{c|}{Edge-level} & \multicolumn{3}{c||}{Recipe-level} & \multicolumn{3}{c}{Average} \\ \cline{3-11}
& & $P$ & $R$ & $F_1$ & $P$ & $R$ & $F_1$ & $P$ & $R$ & $F_1$ \\
\hline \hline
\multirow{4}{*}{Text-Only} & T1. Hand-crafted Features & 57.64 & 54.39 & 55.97 & 59.91 & 57.66 & 58.76 & 58.78 & 56.03 & 57.37 \\
& T2. BERT Pairwise Detector & 67.69 & 83.12 & 74.61 & 76.09 & 85.14 & 78.99 & 71.89 & 84.13 & 76.80  \\
& T3. Ours (Instruction Encoder) & 76.13 & 71.36 & 73.67 & 78.86 & 76.32 & 77.34 & 77.50 & 73.84 & 75.51 \\ 
\hline
\multirow{3}{*}{Image-Only} & I1. Image Similarity Detector & 42.10 & 47.33 & 44.56 & 36.81 & 67.64 & 43.96 & 39.46 & 57.49 & 46.80 \\
& I2. Feed-forward Neural Detector & 35.97 & 47.15 & 40.81 & 50.27 & 56.51 & 50.77 & 43.12 & 51.83 & 45.79 \\
& I3. Ours (Image Encoder) & 57.60 & 59.33 & 58.44 & 65.34 & 68.64 & 68.93 & 61.47 & 63.99 & 62.19 \\ \hline
\multirow{3}{*}{Multi-Modal} & M1. Multi-modal Hand-crafted Features & 60.38 & 58.76 & 59.56 & 62.23 & 60.05 & 61.12 & 61.31 & 59.41 & 60.34 \\
& M2. MMBT Pairwise Detector & 61.86 & \textbf{87.40} & 72.44 & 73.76 & \textbf{88.80} & 78.90  & 67.81  & \textbf{88.10} & 75.67  \\
& M3. Ours (Concatenation)  & 71.20 & 76.32 & 73.67 & 76.26 & 80.79 & 78.50 & 73.73 & 78.56 & 76.09 \\
& M4. Ours (MMBT) & \textbf{77.59} & 79.32 & \textbf{78.45} & \textbf{82.10} & 83.36 & \textbf{82.44} & \textbf{79.84} & 81.34 & \textbf{80.44} \\
\hline 
\end{tabular}
\label{tbl:prerequisite_result}
\end{table*}

\subsection{Performance Comparison}
\label{sec:performance}
Table~\ref{tbl:prerequisite_result} summarize the experimental results comparing against all baseline methods. We analyze the results by answering the following four research questions (RQ). \\

\noindent \textbf{RQ1. Do multi-modal models perform better than ones using a single modality?} Utilizing multi-modal information does achieve better performance in general. M1 outperforms T1 by $2.97$ in average $F$, as M1 has three additional visual-based features. This shows that cooking images provide complementary information to the original text-based feature set of T1. 
Similar results are also observed for neural models. After fusing the textual embeddings of T3 and the visual embeddings of I3 with MMBT, the resultant hybrid M4 model improves over T3 and I3 by $4.93$ and $18.25$, respectively. 

Although utilizing multimodal information is in general beneficial, the means for multimodal fusion is key in taking full advantage of the two modalities. When applying simple feature concatenation (M3) to fuse the two modalities, it only leads to an average $F_1$ gain of $0.58$, compared against the model utilizing only textual information (T3). However, we observe an average $F_1$ gain of $4.35$ when using MMBT (M4) as the fusion method, compared with the method of feature concatenation (M3). We believe this is because the self-attention mechanism in MMBT allows the model to learn the semantic correspondence between cooking images and text descriptions, enabling a more efficient complementing between visual and textual information. \\

\noindent \textbf{RQ2. Which modality is more effective in predicting casual relations -- cooking images or instruction texts?} Although the visual and textual information are complementary, textual information is a necessity in detecting casual relations. Our ablation study shows that the  with instruction-only encoder (T3) outperforms the model with image-only encoder (I3) alone by $13.32$ average $F_1$. By comparing T1 with I1, textual features are also turned out to be more effective than visual features for hand-crafted features. This is inline with our intuition that judging casual relations 
require more deductive reasoning over cooking instructions, rather than inferring intuitively from cooking images. Another possibility is that the image embedding obtained by ResNET do not capture fine-grained visual features (\textit{e.g.}, color, position, or state of certain ingredients) that are crucial for casual relations. \\

\noindent \textbf{RQ3. Do neural-based models perform better than models with hand-crafted features?} Within all three groups of methods (text-only, image-only, and multi-modal), our transformer-based neural models (T3, I3, M4) significantly outperform the methods using hand-crafted features (T1, I1, M1). The average improvements are $19.81$ in precision, $15.41$ in recall, and $17.88$ in $F_1$. For text-only models, the BERT model (T2) outperforms the hand-crafted features (T1) by a large margin. This can be explained by BERT's ability to capture high-level linguistic features such as sentence composition and semantic dependency, which are very important for this task. For image-only models, the feed forward network (I2) achieves comparable results with the model using image similarity (I1), but our model with image encoder (I3) improves I1 by a large margin. This shows that image similarity already serves as a strong clue in determining pairwise casual relations, but the model is more effective when it gets access to all the cooking images in the recipe. 




\section{Conclusion}
\label{sec:conclusion}

We investigate the problem of automatically building cooking workflows for food recipes, leveraging both text and images. We build the Multi-Modal Recipe Structure dataset (MM-ReS), the first large-scale dataset for this task, containing 9,850 food recipes with labeled cooking workflows. We also present a novel encoder--decoder framework, which applies Multimodal Bitransformers (MMBT) to fuse visual and textual features.  Our solution couples the use of the transformer model and the pointer network to utilize the entire recipe context. Experimental results on MM-ReS show that considering multimodal features enables better performance for detecting causal relations, and the cooking images are highly complementary to procedure text. 

\begin{acks}
This research is supported by the National Research Foundation, Singapore under its International Research Centres in Singapore Funding Initiative. Any opinions, findings and conclusions or recommendations expressed in this material are those of the author(s) and do not reflect the views of National Research Foundation, Singapore. 
\end{acks}

\bibliographystyle{ACM-Reference-Format}
\bibliography{sample-base}


\begin{figure*}[!t]
\centering
\caption{The cooking workflow of ``Eggplant Parmesan'' (a), and the predicted workflows for different methods (b-d). }
\subfigure[Correct Cooking Workflow]
{
	\begin{minipage}[t]{0.47\linewidth}
	\centering
	\includegraphics[width=5.8cm]{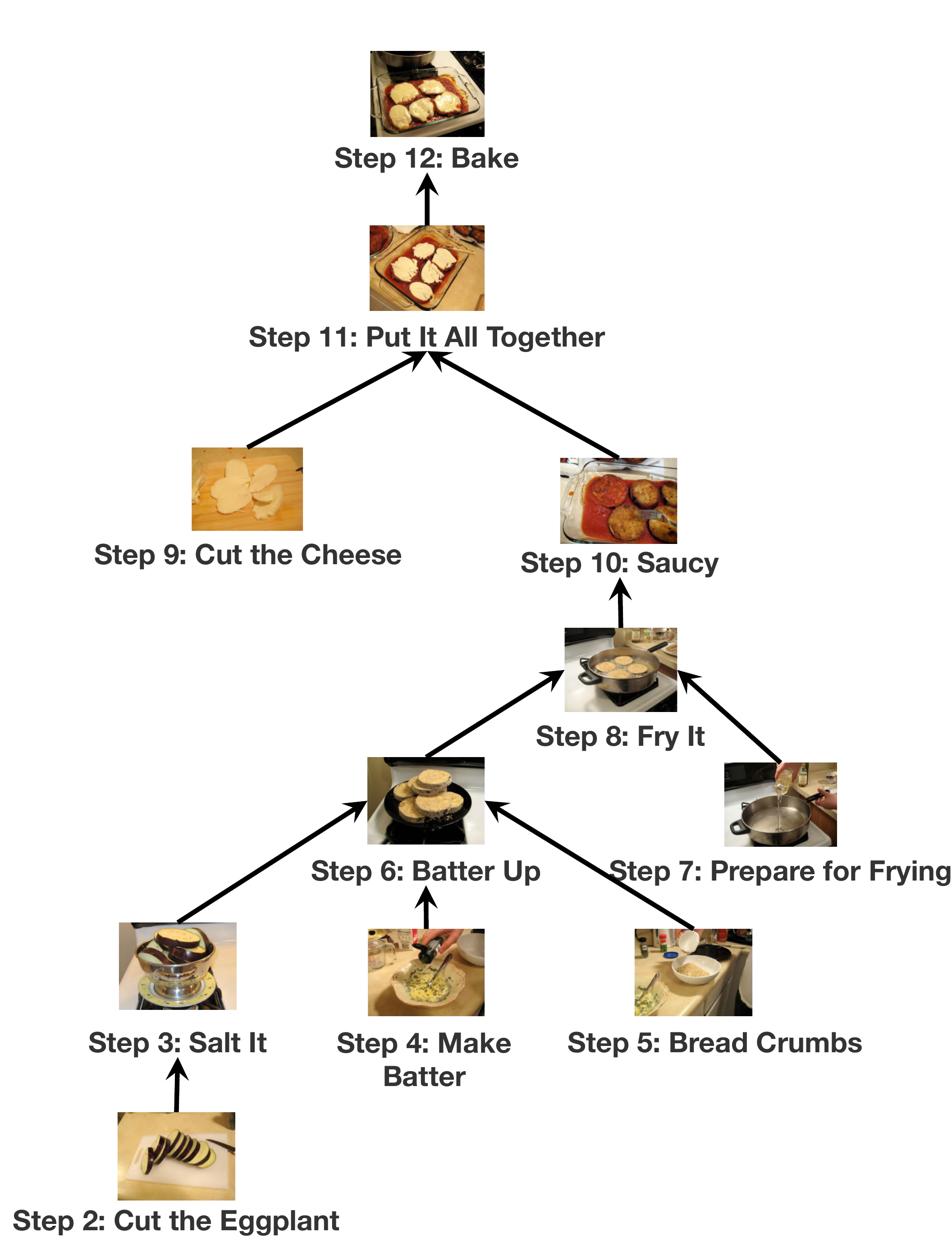}
	\end{minipage}
	\label{fig:eggplant_truth}
}
\subfigure[Ours (MMBT)]
{
	\begin{minipage}[t]{0.47\linewidth}
	\centering
	\includegraphics[width=5.8cm]{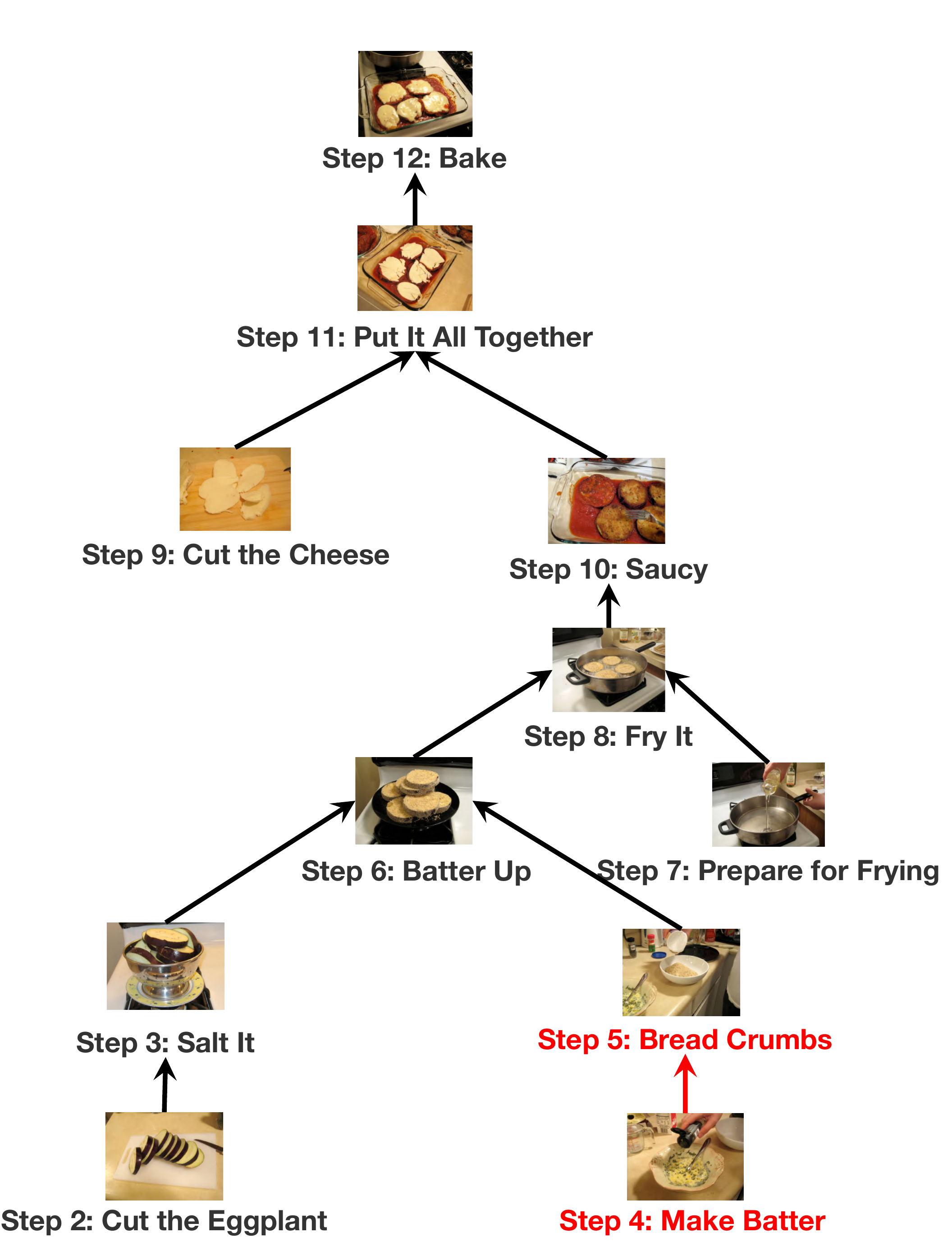}
	\end{minipage}
	\label{fig:eggplant_multi}
}
\subfigure[Ours (Instruction Encoder)]
{
	\begin{minipage}[t]{0.47\linewidth}
	\centering
	\includegraphics[width=5.8cm]{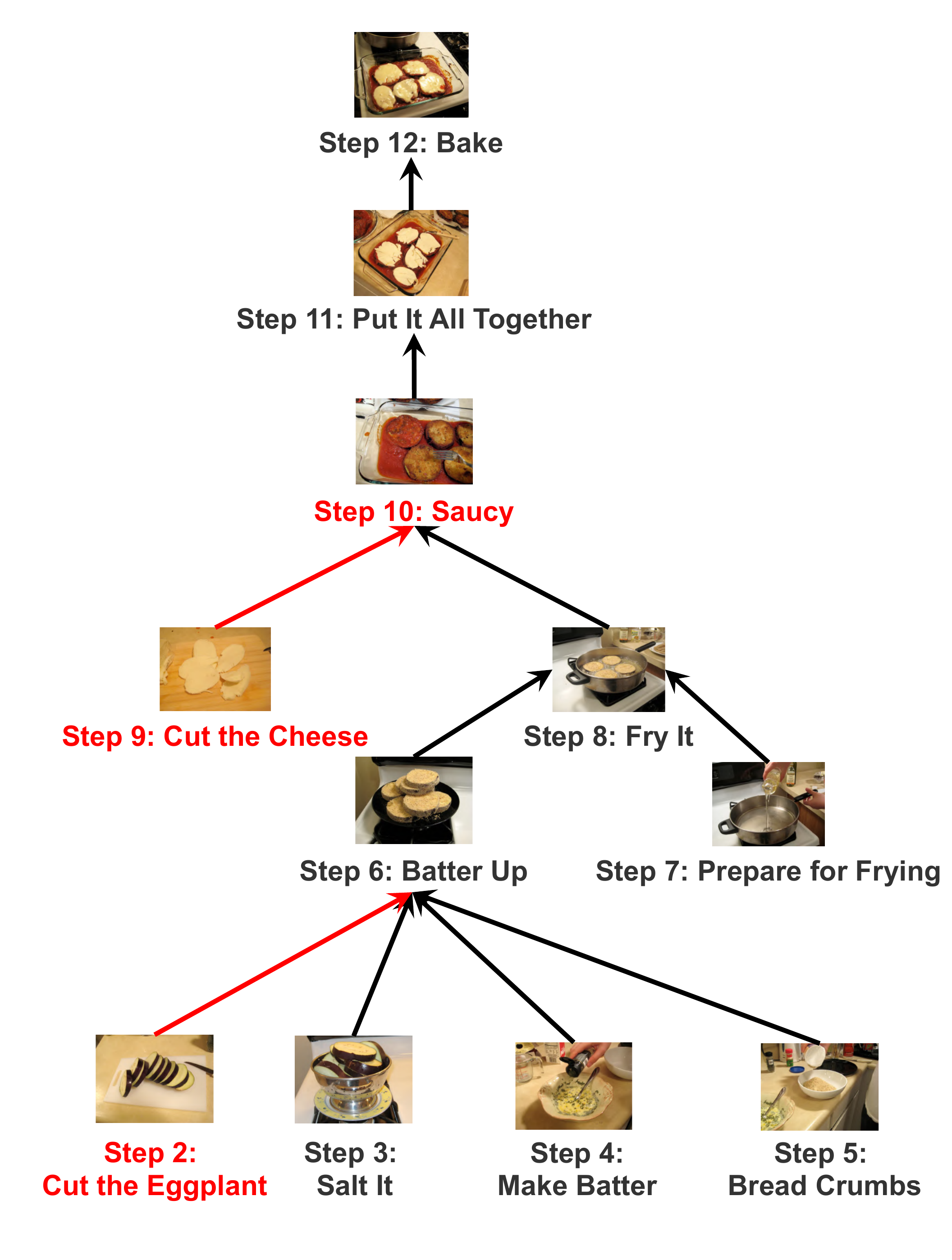}
	\end{minipage}
	\label{fig:eggplant_text}
}
\subfigure[Ours (Image Encoder)]
{
	\begin{minipage}[t]{0.47\linewidth}
	\centering
	\includegraphics[width=5.8cm]{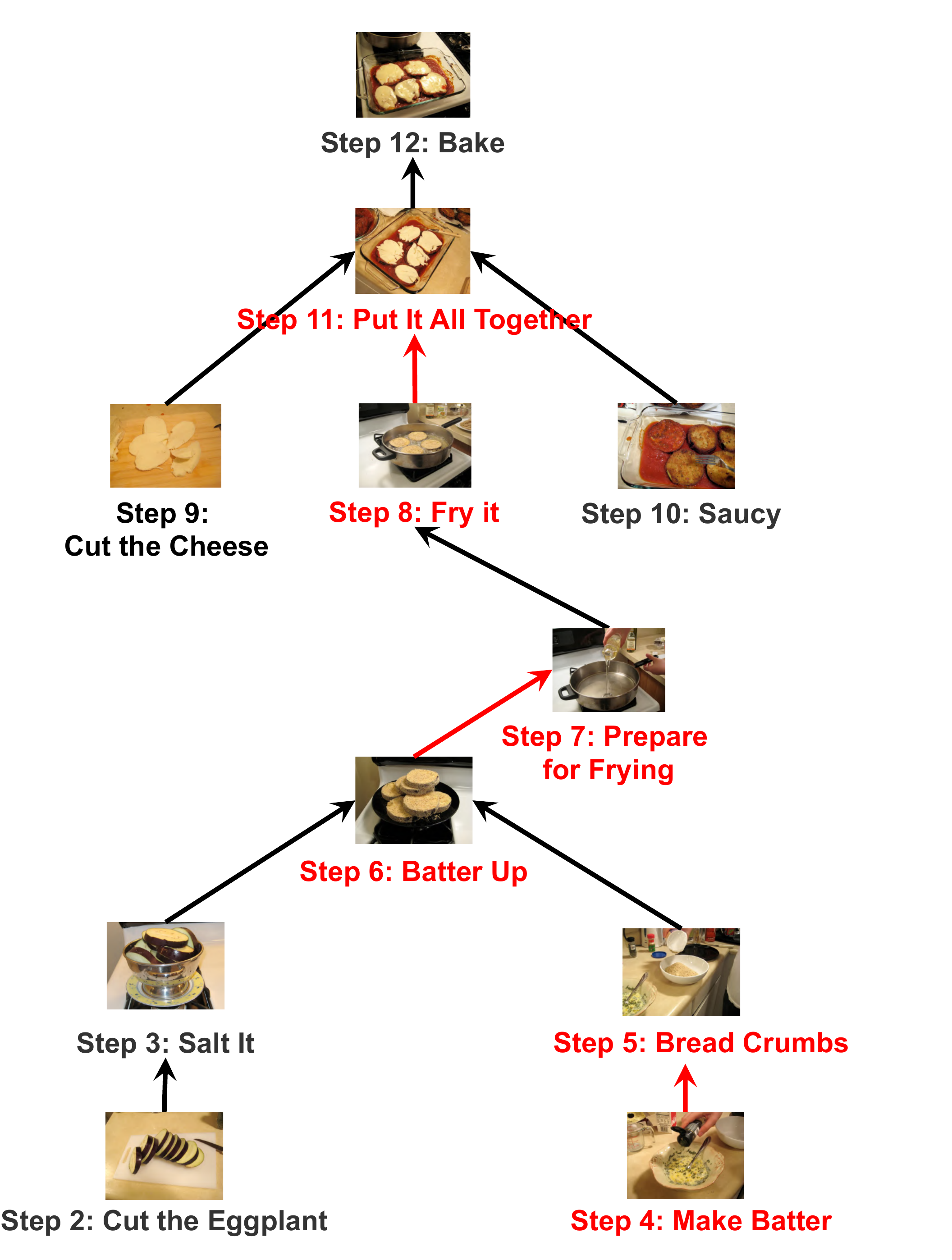}
	\end{minipage}
	\label{fig:eggplant_image}
}
\label{fig:tree_visualization}
\end{figure*}

\appendix

\section{Result Visualization and Error Analysis}
\label{sec:case_study}

To intuitively understand the effectiveness of multi-modal fusion, we select the recipe of ``Eggplant Parmesan'' for case study in Figure~\ref{fig:tree_visualization}. The linking errors made by each method are highlighted in red. Three errors are made by the model when we only utilize the cooking image for casual relation detection (I3). The model tend to mistakenly treat two steps as sequential if their cooking images are visually similar. For example, Step 4 and Step 5 are in parallel but are predicted as sequential since they look similar visually. The same mistake happens between Step 6 and Step 7. The text-only model (T3) correct the above two errors, but mistakenly treat Step 2 and Step 3 as parallel, as shown in Figure~\ref{fig:eggplant_text}. The sequential relation between these two steps are easy to judge visually, but hard to tell from text description because Step 3 does not mention ``Eggplant'' in contexts. When fusing both visual and textual information, our multi-modal model (M4) gets the best result, making only one error in this example (the relation between Step 4 and Step 5). 

\end{document}